\documentclass[12pt]{article} 

\usepackage{algorithm}
\usepackage{algorithmic}
\usepackage{subcaption}
\usepackage{tabularx}
\usepackage{adjustbox}
\usepackage{float}
\usepackage{placeins}
\usepackage{booktabs}

\usepackage{hyperref}
\usepackage{amsmath}
\usepackage{amsthm}
\usepackage{amssymb}
\usepackage{breqn}
\usepackage{algorithm}
\usepackage{algorithmic}

\newcommand*\samethanks[1][\value{footnote}]{\footnotemark[#1]}

\begin{document}
	\title{Gaussian Hierarchical Latent Dirichlet Allocation: Bringing Polysemy Back}	
	\author{
		Takahiro Yoshida\thanks{Equal contribution} \\
		Graduate School of Information Science and Technology \\
		The University of Tokyo\\
		\and
		Ryohei Hisano\samethanks[1] \\
		Graduate School of Information Science and Technology \\
		The University of Tokyo\\
		\and
		Takaaki Ohnishi\\
		Graduate School of Information Science and Technology \\
		The University of Tokyo\\
	}
	\maketitle
	\thispagestyle{empty}
	
\begin{abstract}

Topic models are widely used to discover the latent representation of a set of documents. The two canonical models are latent Dirichlet allocation, and Gaussian latent Dirichlet allocation, where the former uses multinomial distributions over words, and the latter uses multivariate Gaussian distributions over pre-trained word embedding vectors as the latent topic representations, respectively. Compared with latent Dirichlet allocation, Gaussian latent Dirichlet allocation is limited in the sense that it does not capture the polysemy of a word such as ``bank.''  In this paper, we show that Gaussian latent Dirichlet allocation could recover the ability to capture polysemy by introducing a hierarchical structure in the set of topics that the model can use to represent a given document. Our Gaussian hierarchical latent Dirichlet allocation significantly improves polysemy detection compared with Gaussian-based models and provides more parsimonious topic representations compared with hierarchical latent Dirichlet allocation. Our extensive quantitative experiments show that our model also achieves better topic coherence and held-out document predictive accuracy over a wide range of corpus and word embedding vectors.

\end{abstract}

\section{Introduction}

% Topic models in genearal
Topic models are widely used to identify the latent representation of a set of documents.  Since latent Dirichlet allocation (LDA)~\cite{Blei2003} was introduced, topic models have been used in a wide variety of applications.  Recent work includes the analysis of legislative text~\cite{ONeill2016}, detection of malicious websites~\cite{Wen2018}, and analysis of the narratives of dermatological disease~\cite{Obot2018}.  The modular structure of LDA, and graphical models in general~\cite{Lauritzen1996}, has made it possible to create various extensions to the plain vanilla version. Significant works include the correlated topic model (CTM), which incorporates the correlation among topics that co-occur in a document~\cite{Blei2005}; hierarchical LDA (hLDA), which jointly learns the underlying topic and the hierarchical relational structure among topics~\cite{Blei2010}; and the dynamic topic model, which models the time evolution of topics~\cite{Blei2006}.

%, and relational topic model that adds relational structure among documents (e.g., cite information)~\cite{Chang2009} among others.

% GLDA
LDA uses multinomial distributions over words, whereas Gaussian LDA (GLDA)~\cite{Das2015} uses multivariate Gaussian distributions over a pre-trained word embedding to represent the underlying topics.  Using the word embedding vector space representation, GLDA has the added benefit of incorporating semantic regularities in a language, which results in increasing coherency~\cite{Newman2010, TopicEval2015,ChangReadTea} of topics~\cite{Das2015}.  Recent developments of this line of research include correlated Gaussian topic models (CGTM)~\cite{Xun2017}, which add a correlational structure to the topics used in a document; the work of \cite{Batmanghelich2016}, which replaces the Gaussian distribution with a von Mises--Fisher distribution; and the latent concept topic model~\cite{Tsujii2017}, which redefines each topic as the distribution over latent concepts, where the latent concept is modeled as a multivariate Gaussian distribution over the word embeddings.

A crucial discrepancy of GLDA and CGTM is that they fail to detect the polysemy of a term, such as ``bank,'' which LDA and hLDA capture well~\cite{Steyvers2006}. LDA is a mixed membership model with no mutual exclusivity constraint that restricts the assignment of words to one topic only~\cite{Steyvers2006}. As we show in the current paper, the delicate balance between a term that captures the probability of a word under a topic, and the probability of a topic given a document, in the collapsed Gibbs sampler of LDA~\cite{Griffiths2004}, makes it possible to capture polysemy. However, although GLDA and CGTM are mixed membership models with no mutual exclusivity constraint, the probability of a word under a topic is characterized by a multivariate $\mathcal{T}$ distribution that outweighs the term that reflects the likelihood of a topic given a document. Hence, mutual exclusivity is likely to be unintentionally recovered, and the ability to detect polysemy is lost.

% So what we do
In this paper, we show that the ability to capture polysemy in GLDA-type models can be recovered by restricting the set of topics that can be used to represent a given document.  One parsimonious implementation of such a restriction can be achieved by incorporating a hierarchical topic structure, as in hLDA~\cite{Blei2010, Blei2003HLDA}.  In our Gaussian hLDA, topics that can be used in a document are restricted by a path of topics that are learned jointly from the data.  Instead of assigning a topic to each word position in a document, we assign levels that describe the position of the path from which the word was sampled.

At first glance, our model may seem to have a price to pay in terms of time complexity because of the added complexity of the model.  However, because we do not need to sample from the entire set of topics for each word position in a document, the time complexity of our model does not necessarily worsen compared with GLDA and CGTM.  Moreover, our model has the benefit of capturing polysemy in addition to being able to learn a compact hierarchical structure that shows the relationships among topics. Additionally, as in hLDA~\cite{Blei2010}, Bayesian nonparametric techniques can also be used, thereby making it possible to determine the hierarchical tree structure more flexibly.

% Other related literature: Topic and wordembedding
Other works also exist that combine topic modeling and word embeddings.  \cite{Petterson2010} used information from the word similarity graph to achieve more coherent topics.  \cite{Nguyen2015} modified the likelihood of the model by combining information from pre-trained word embeddings with a log-linear function. Instead of using pre-trained word embedding vectors, some works have attempted to learn word embeddings and topics from the corpus jointly.  The embedded topic model\cite{Dieng2019b} uses the inner product between a word embedding and topic embedding as the natural parameter that governs the multinomial distribution and learns the two representations from the corpus simultaneously.  In \cite{Dieng2019}, the model was further extended to incorporate the time evolution of the topic embeddings.  The Wasserstein topic model\cite{Xu2018} unifies topic modeling and word embedding using the framework of Wasserstein learning. Compared with these models, we leave the word embedding vectors as it is and enrich the topic co-occurrence structure of a document to adapt to the corpus of interest.

% DL and word embdding in genearal
%Another strand of research is the deep learning approaches.

\begin{figure}
	\centering
	\includegraphics[width=0.6\linewidth]{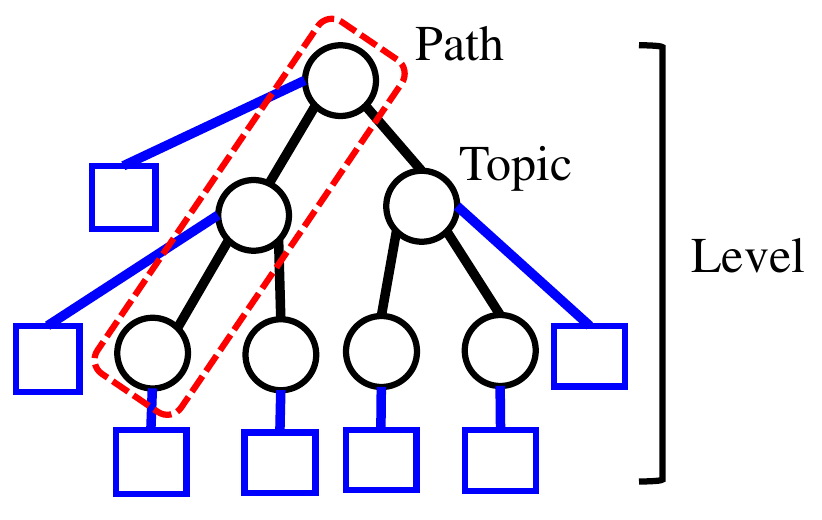}
	\caption{Hierarchy among topics: a circle represents a topic, dashed rectangle represents a path, each layer represents a level, and  rectangular nodes represent the candidate branch that might be added by the nCRP.}
	\label{fig:hierarchy}
\end{figure}

% Contribution
Our contributions are summarized as follows:

\begin{itemize}
	\item We propose the Gaussian hLDA, which significantly improves the capture of polysemy compared with GLDA and CGTM.
	\item Our model jointly learns the topics, in addition to the hierarchical structure, and characterizes the relationship among the topics.  The hierarchical structure can also be used to analyze the correlation structure among topics.
	\item The hierarchical tree structure can be estimated in a flexible manner using the nested Chinese restaurant process~\cite{Blei2010}.
	\item Even though our model is far more expressive than GLDA and CGTM, the time complexity does not necessarily worsen compared with that of those two models.
	\item We show that our model exhibits a more parsimonious representation of topics than hLDA.
	\item Using three real-world corpora and three different pre-trained word embedding vectors, we show that our model outperforms state-of-the-art models both in terms of the held-out predictive likelihood and topic coherence.
\end{itemize}

\section{Notation}
We briefly summarize the mathematical notation used throughout the paper.  $D$ denotes the number of documents in a corpus,  $V$ denotes the number of unique words in the corpus, $K$ denotes the number of topics, $M$ denotes the dimension of the word embedding vector, and $L$ denotes the depth of the maximum level of the hierarchy.  Lower-case letters (e.g., $d$, $v$, and $k$) denote a specific document, word, or topic.  $\theta_{d,k}$ denotes the probability of topic $k$ for document $d$ and, $\phi_{k,v}$ denotes the probability of word $v$ in topic $k$.  $N_d$ denotes the number of words in a document. For each word position $n$ in document $d$, $z_{d,n}$ denotes either the topic or level assignment for that word position and $w_{d,n}$ denotes the word that appears in word position $n$ for document $d$.  Furthermore, $c_{d}$ denotes a path assignment to document $d$ and $l_{d}$ denotes the level distribution in $d$.  For path and level assignments, a topic is uniquely defined as shown in Fig.~\ref{fig:hierarchy}.  $N_{d(-n)}^{i}$ denotes the number of word positions in $d$ that are assigned to either topic (LDA, GLDA, CGTM) or level $i$ (hLDA, GhLDA), excluding $z_{d,n}$. $N_{d(-n)}^{>i}$ is defined similarly counting the number of word positions above level $i$. $N_{-(d,n)}^{kv}$ denotes the number of word positions in the entire corpus with word $v$ and topic $k$, excluding $z_{d,n}$.
$GEM(m,b)$ denotes the Griffiths, Engen, and McCloskey distribution~\cite{Pitman2002}, which is used to define a prior level distribution among a path, and $m$ and $b$ denote hyperparameters that control the stick-breaking process.  $nCRP(\gamma)$ represents the nested Chinese restaurant process ~\cite{Blei2010}, where $\gamma$ denotes a hyperparameter that controls the probability of a new branch that emerges in the current tree (i.e., the parameter that controls the likelihood of the blue rectangle being chosen in Fig.~\ref{fig:hierarchy}).	
$Dir(\alpha)$ represents a Dirichlet distribution and $Mult$ represents a multinomial distribution, where $\alpha$ denotes a hyperparameter vector.  $N(\mu,\Sigma)$ and $\mathcal{T}_{v}(\mu,\Sigma)$ denote a normal distribution and multivariate $\mathcal{T}$ distribution with mean vector $\mu$ and covariance matrix $\Sigma$, respectively.  $\mathcal{NIW}(u,\Psi,v,\kappa)$ denotes a normal inverse Wishart distribution with hyperparameters $u,\Psi,v,\kappa$, where $u$ denotes a vector, $\Psi$ denotes a matrix, and $v$ and $\kappa$ denote positive real values.  Furthermore, $\kappa_{k} = \kappa + |s_{k}| $, $v_{k}= v + |s_{k}|$, $\Psi_{s_{k}} = \Psi + \frac{\kappa |s_{k}|}{\kappa_{k}} (\bar{x}_{s_{k}}-u)(\bar{x}_{s_{k}}-u)^{T} + \Sigma_{i \in s_{k}} (x_{i} - \bar{x}_{s_{k}})(x_{i} -\bar{x}_{s_{k}})^{T}$, $u_{k}=\frac{\kappa\mu_{0} + |s_{k}|\bar{x}_{s_{k}}}{\kappa_{k}}$, where $s_{k}$ denotes the set of indicators of word positions that is assigned to topic $k$ and $\bar{x}_{s_{k}}$ denotes the mean vector among the indicators in $s_{k}$.

\section{Related Work}
\subsection{Gaussian Latent Dirichlet Allocation}

The generative process of LDA and GLDA can be written similarly, and we focus on the GLDA case.  GLDA uses word embedding vectors to characterize words in a document.  We define $D, N_d, K, \theta_d, z_{d,n}$ precisely, as summarized in the previous section.  Instead of considering $w_{d,n}$ as an indicator that denotes a word, as in LDA, we consider it as a vector from a pre-trained word embedding.  The generative process is summarized as follows:

\begin{tabbing}
	\hspace{10pt}\=\hspace{10pt}\=\kill
	(1) For all topics $k$, sample $\mu_k,\Sigma_k \sim \mathcal{NIW}(u,\Psi,\kappa,v)$. \\
	(2) For each document $d$, \\
	\> (a) sample topic proportion $\theta_{d} \sim Dir(\alpha)$; and \\
	\> (b) for each word position in $d$, sample topic \\
	\>\> assignments $z_{d,n} \sim Mult(\theta_{d})$ and words from \\
	\>\> $w_{d,n} \sim \mathcal{N}(\mu_{z_{d,n}},\Sigma_{z_{d,n}})$.
\end{tabbing}

The collapsed Gibbs sampler of GLDA can be written as
\begin{align}
	\label{glda-sampler}
	\begin{split}
		& p(z_{d,n}=k|w,z_{-(d,n)})\propto  \frac{\alpha_k+N_{d(-n)}^{k}}{\sum_{k'}(\alpha_{k'}+N_{d(-n)}^{k'})}\cdot  \\
		&\mathcal{T}_{v_k-M+1}\left(w_{d,n}|u_k,\frac{\kappa_k+1}{\kappa_k(v_k-M+1)}\Psi_{s_{k}}\right).
	\end{split}
\end{align}
\noindent LDA is recovered by replacing ``$\mu_k,\Sigma_k \sim \mathcal{NIW}(u,\Psi,\kappa,v)$'' in (1) with ``$\phi_{k} \sim Dir(\beta)$,'' ``$w_{d,n} \sim \mathcal{N}(\mu_{z_{d,n}},\Sigma_{z_{d,n}})$'' in (2)(b) with ``$w_{d,n} \sim Mult(\phi_{z_{d,n}})$,'' and the second term in the sampler with ``$\frac{\beta_v+N_{-(dn)}^{kv}}{\sum_{v'}(\beta_{v'}+N_{-(dn)}^{kv'} )}$.''

\subsection{Correlated Gaussian Topic Model}
CGTM~\cite{Xun2017} is an extension of GLDA that incorporates correlation among topics used in a document, similar to CTM~\cite{Blei2005}.  The generative process is summarized as follows:

\begin{tabbing}
	\hspace{10pt}\=\hspace{10pt}\=\kill
	(1) For all topics $k$, sample $\mu_k,\Sigma_k \sim \mathcal{NIW}(u,\Psi,\kappa,v)$. \\
	(2) To model the correlation among topics, sample \\
	\> $\mu_a,\Sigma_a \sim \mathcal{NIW}(u_a,\Psi_a,\kappa_a,v_a)$. \\
	(3) For all documents $d$, \\
	\> (a) sample $\eta_d \sim \mathcal{N}(\mu_a,\Sigma_a)$; \\
	\> (b) transform $\eta_{d}$ to a topic proportion vector $\theta_{d}$ \\
	\>\> using a softmax function $\theta_d=\frac{\exp(\eta_d)}{\sum_k \exp(\eta_{d,k})}$; and \\
	\> (c) for all word positions in $d$, sample topic assignments  \\
	\>\>  $z_{d,n} \sim Mult(\theta_{d})$ and resulting words from \\
	\>\> $w_{d,n} \sim \mathcal{N}(\mu_{z_{d,n}},\Sigma_{z_{d,n}})$.
\end{tabbing}

CGTM can be estimated by alternatively sampling $\eta_{d}$  and topic assignments for each word position $z_{d,n}$.  The sampling of $\eta_{d}$ is rather involved, and includes additional auxiliary variable $\lambda_{d}$ and sampling from a Polya--Gamma distribution\cite{Polson2012,Makalic2016}.  After $\eta_{d}$ (and therefore $\theta_{d}$) is sampled, the topic assignments $z_{d,n}$s are sampled using
\begin{equation}
\label{cgtm-sampler}
\begin{split}
& p(z_{d,n}=k|w,z_{-(d,n)})\propto \frac{exp(\eta_{d}^{k})}{\Sigma_{i}exp(\eta_{d}^{i})} \cdot \\
& \mathcal{T}_{v_k-M+1}\left(w_{d,n}|u_k,\frac{\kappa_k+1}{\kappa_k(v_k-M+1)}\Psi_{s_{k}}\right).
\end{split}
\end{equation}
\subsection{Hierarchical Latent Dirichlet Allocation}

The goal of hLDA is to identify topics and hierarchical relationships among the topics simultaneously from the corpus. Words in a document are drawn from the restricted set of topics that are characterized using paths from the hierarchical topic structure.  Because of the hierarchical tree structure, topics in the upper level are used more frequently and thus capture more general terms than the lower level.  To learn the hierarchical structure more flexibly, hLDA\cite{Blei2010} uses the nested Chinese restaurant process as the prior distribution that defines the hierarchy over topics.  The generative process is summarized as follows:

\begin{tabbing}
	\hspace{5pt}\=\hspace{5pt}\=\kill
	(1) For all topics $k$, sample $\phi_{k} \sim Dir(\beta)$. \\
	(2) For each document $d$, \\
	\> (a) sample a path assignment $c_{d} \sim nCRP(\gamma)$; \\
	\> (b) sample a distribution over levels in the path, \\
	\>\> $l_{d} \sim GEM(m,b)$; and \\
	\> (c) for all word positions in $d$, first choose the level \\
	\>\> assignments $z_{d,n} \sim Mult(l_{d})$ and then the resulting  \\
	\>\> words from the topic at that level in the path, $w_{d,n} \sim$ \\
	\>\> $Mult(\phi_{c_{d}[z_{d,n}]})$.
\end{tabbing}

In hLDA, we need to sample both the path assignments for all documents and level assignments for all word positions.  The Gibbs sampling algorithm is similar to those used in GhLDA, so we omit it here.

\section{Gaussian Hierarchical Latent Dirichlet Allocation}

\subsection{Mutual Exclusivity}\label{praticallyGMM}

The problem with GLDA and CGTM can be clarified by considering the sampling equations of GLDA (i.e., Eq.~\ref{glda-sampler}) and CGTM (i.e., Eq.~\ref{cgtm-sampler}).  Two observations are worth mentioning.  First, the only difference between  Eq.\ref{glda-sampler} and Eq.\ref{cgtm-sampler} is the first term on the right-hand side of each equation, which corresponds to the probability of a topic given a document (Eq.~\ref{glda-sampler}) and the probability of a topic given a document with correlation (Eq.~\ref{cgtm-sampler}).

Second, although the first term on the right-hand side of the sampling equation can vary at most in the order of $O(10^{-N_{d}})$ among the topics, the second term is a multivariate probability density function that can vary much more widely.  The order of variability of the $\mathcal{T}$ distribution among the topics widens when the data points in the word embedding that we want to cluster are multimodal, thereby ensuring each centroid of the Gaussian mixture to be placed in distinct positions in the word embedding space.  Similar words in a word embedding space tend to cluster together, which makes word embeddings far from unimodal.  This condition results in the second term outweighing the first term, and mutual exclusivity is likely to be unintentionally recovered in GLDA and CGTM.

\subsection{Gaussian Hierarchical Latent Dirichlet Allocation}

To create a mixed membership model with no mutual exclusivity constraint, even in cases that consider multivariate Gaussian distributions, we need to go beyond merely sampling topic assignments for each word position in the corpus and restrict the set of topics that can be used to represent a given document.  By doing so, when a topic such as ``finance, bank, loan'' appears in a document, we can only use a particular topic such as ``banks, ratio, interest'' without being able to sample from all the available topics.  This restriction guarantees that there is no restriction on mutual exclusivity and, as a bonus, can be used to capture the correlation among topics.  One straightforward approach to add this constraint is via hierarchical topic modeling, as in \cite{Blei2010, Blei2003HLDA}.  In the hierarchical construction, topics are ordered according to the level of abstraction from top to bottom.  Path $c_{d}$ is used to characterize the topics that can be used in a document $d$, and each word position in a document has level assignments $l_{d,n}$s that capture the level at which the word is sampled.

The generating process of GhLDA is as follows:
\begin{tabbing}
	\hspace{5pt}\=\hspace{5pt}\=\kill
	(1) For all topics $k$, sample $\mu_k,\Sigma_k \sim \mathcal{NIW}(u,\Psi,\kappa,v)$.\\
	(2) For each document $d$, \\
	\> (a) sample a path assignment $c_{d} \sim nCRP(\gamma)$; \\
	\> (b) sample a distribution over level of the path: \\
	\>\> $l_{d} \sim GEM(m,b)$; and \\
	\> (c) for all word positions in $d$, first choose the level \\
	\>\> assignments $z_{d,n} \sim Mult(l_{d})$ and the resulting words from \\
	\>\> the topic at level $z_{d,n}$ in the path, $w_{d,n} \sim \mathcal{N}(\mu_{c_{d}[z_{d,n}]},$ \\
	\>\> $\Sigma_{c_{d}[z_{d,n}]})$.
\end{tabbing}

\begin{table}
	\caption{Summary of qualitative characteristics}
	\label{characteristic}
	\centering
	\resizebox{0.7\columnwidth}{!}{
		\begin{tabular}{lllll}
			\toprule
			Model & Pruning  & Polysemy & Correlation & Embedding  \\
			\midrule
			LDA &  $\times$ &  $\bigcirc$  &  $\times$ & $\times$\\
			hLDA &  $\bigcirc$ &  $\bigcirc$ &  $\bigcirc$ & $\times$\\
			GLDA &  $\bigcirc$ &  $\times$ &  $\times$ & $\bigcirc$\\
			CGTM &  $\bigcirc$ &   $\times$ &  $\bigcirc$ &  $\bigcirc$\\
			GhLDA &  $\bigcirc$ &  $\bigcirc$ &  $\bigcirc$ & $\bigcirc$\\
			\bottomrule
	\end{tabular}}
\end{table}

\subsection{Gibbs Sampling Algorithm}

We need to sample both the path assignments for all documents $d$ and level assignments for all word positions $w_{d,n}$.  The Gibbs sampling algorithm is as follows;
\begin{tabbing}
	\hspace{5pt}\=\hspace{5pt}\=\kill
	(1) For each document $d$, first sample path assignment \\
	\>\> $c_{d} \sim p(c_{d}|w,c_{-d},z,H) p(w_{d}|c,w_{-d},z,H)$; and \\
	(2) for all word positions in $d$, sample level assignments \\
	\>\> $p(z_{d,n}|z_{-(d,n)},c,w,H)\propto p(z_{d,n}|z_{d,-n},H)$\\
	\>\> $p(w_{d,n}|z,c,w_{-(d,n)},H)$,
\end{tabbing}
\noindent where $H$ is the set of hyperparameters in the model.  The probability of a path is the product of the prior on paths defined by $nCRP(\gamma)$ (i.e., $p(c_{d}|w,c_{-d},z,H)$)~\cite{Blei2010}, and the probability of a word given a specific path, which is

\begin{align}
	\begin{split}
		&p(w_{d}|c,w_{-d},z,H) = \prod_{l=1}^L \frac{1}{\pi^{tM/2}}\frac{\Gamma_{M}(\frac{v+|s_{c[l]}|+|t_{l}|}{2})}{\Gamma_{M}(\frac{v+|s_{c[l]}|}{2})}\cdot\\
		&\frac{|\Psi_{|s_{c[l]}|}|^{\frac{v+|s_{c[l]}|}{2}}}{|\Psi_{|s_{c[l]}|+|t_{l}|}|^{\frac{v+|s_{c[l]}|+|t_{l}|}{2}}}(\frac{\kappa+|s_{c[l]}|}{\kappa+|s_{c[l]}|+|t_{l}|})^{M/2},
	\end{split}
	\label{BNPpathsample}
\end{align}

\noindent where $N_{-d}^{c}$ denotes the number of documents assigned to path $c$, excluding $d$, $s_{c[l]}$ denotes the set of word positions assigned to topic $c[l]$, $t_{l}$ denotes the set of word positions assigned to level $l$ in $d$, and $\Gamma_{d}$ denotes the multivariate gamma function.  The probability of a level is defined as
\begin{align}
	\begin{split}
		&p(z_{d,n}=l|c,z_{-(d,n)},w)=\\
		&\frac{mb+N_{d(-n)}^{l}}{b+N_{d(-n)}^{\geq l}}\prod_{i=1}^{i=l-1}\frac{(1-m)b+N_{d(-n)}^{>l}}{b+N_{d(-n)}^{\geq l}}\cdot\\
		&\mathcal{T}_{v_k-M+1}\left(w_{d,n}|u_k,\frac{\kappa_k+1}{\kappa_k(v_k-M+1)}\Psi_{s_{c[l]}}\right).
	\end{split}
\end{align}
The qualitative characteristics of LDA, hLDA, GLDA, CGTM, and GhLDA are summarized in Table~\ref{characteristic}.  Pruning implies the necessity to prune highly frequent words, such as stop words, from the corpus.  Whereas LDA fails to provide interpretable topics without pruning, all the other models handle this with ease.   Polysemy implies the ability to capture polysemy.  The manner in which GLDA and CGTM fail is described in Section 5.  Correlation implies capturing the co-occurrence of topics in a document and embedding means the use of pre-trained word embedding vectors.

\subsection{Complexity Analysis}
\begin{table}[t]
	\caption{Running time complexities}
	\label{time-complexity}
	\vskip 0.15in
	\begin{center}
		\begin{small}
			%\begin{sc}
			\begin{tabular}{lc}
				\toprule
				Model & Complexity \\
				\midrule
				LDA  & $O( N_{d}K)$ \\
				hLDA & $O(K+ N_{d} L)$ \\
				GLDA & $O(N_{d}KM^2)$ \\
				CGTM & $O(K^3 + N_{d}KM^2)$ \\
				GhLDA & $O(KM^2 + N_{d}LM^2)$ \\
				\bottomrule
			\end{tabular}
			%\end{sc}
		\end{small}
	\end{center}
	\vskip -0.1in
\end{table}

We compare the running time complexity of all the models. Because hLDA, CGTM, and GhLDA include steps that require us to sample document-level parameters using all the words that appear in a document, we focus on the running time complexity to sample all assignments for a given document $d$.  Table~\ref{time-complexity} summarizes the time complexities.  Each sampling step in GLDA requires us to evaluate the determinant and inverse of the posterior covariance matrix, which is cubic.  However, as indicated by \cite{Das2015}, this can be reduced to $O(M^{2})$ using the Cholesky decomposition of a covariance matrix.  Because each word position has $K$ topics to consider, and there are $N_{d}$ words in a document, the total time complexity of GLDA is $O(N_{d}KM^2)$.  LDA does not require us to calculate the inverse of the posterior covariance matrix, which makes the time complexity $O(N_{d}K)$.  For each document, CGTM requires the sampling of document-level parameters $\eta_{d}$ and $\lambda_{d}$.  This step adds another $O(K^3)$ to the complexity.

Compared with these models, GhLDA first evaluates the posterior predictive probability for all paths.  The straightforward calculation results in $O(PLM^{2})$, where $P$ denotes the number of paths and $L$ denotes the maximum depth among all paths. However, exploiting the tree structure, we can reduce the calculation to $O(KM^2)$.  After sampling the path, GhLDA proceeds to sample levels for each word position in a document.  Because each path only has a most $L$ topics, sampling-level assignment for all words in a document takes $O(N_{d}LM^2)$.  Adding both steps leads to $O(KM^2 + N_{d}LM^2)$ in total.  Similar arguments can be used to calculate the time complexity of hLDA, which is $O(K+ N_{d} L)$.

A few points are worth mentioning.  All the models that use word embedding vectors are much slower than their plain counterparts because of the additional step of computing the Cholesky decomposition.  However, comparing GLDA and GhLDA, we can see that GhLDA does not necessarily increase the time complexity compared with GLDA.  If $N_{d}K \leq K+N_{d}L$, the time complexity of GhLDA is lower than that of GLDA\footnote{This is indeed a reasonable scenario.  For instance, assume that there are 100 words in a document $d$ (i.e., $N_{d}=100$). Whereas GLDA with $K=20$ leads to $N_{d} \times K = 2,000$, GhLDA with the branch structure of $[1,1,4,4]$ (i.e., $K=22$ and $L=4$) results in $K+N_{d}L=422$.}.  Surely enough, this argument does not take into account the number of iterations required for collapsed Gibbs sampling to converge.  However, it still highlights the fact that the time complexity of GhLDA is not necessarily worse than that of GLDA.

\begin{table}
	\caption{Selected topics related to ``Rivers'' and ``Banks/Financial'' in the Wikipedia dataset}
	\label{Table: GLDA}
	\centering
	\scalebox{0.9}{
		\begin{tabular}{cl}
			\toprule
			Model  & Topic and Top 5 Words \\
			\midrule
			LDA & 0 [the,in,creek,is,it]\\
			(K=20) & 2 [the,in,is,financial,for]\\\hline
			GLDA & 10 [bank, financial,banks,banking,central]\\
			(K=20) & 13 [creek,de,lake,water,french]\\\hline
			GLDA & 0 [river,bank,creek,flows,group]\\
			(K=40) & 21 [police,financial,banking,market,management]\\\hline
			CGTM & 14 [bank,financial,university,mathematical,theory]\\
			(K=20) & 17 [police,creek,air,services,lake]\\
			\bottomrule
	\end{tabular}}
\end{table}

\begin{figure}
		\centering
		\includegraphics[width=0.5\linewidth]{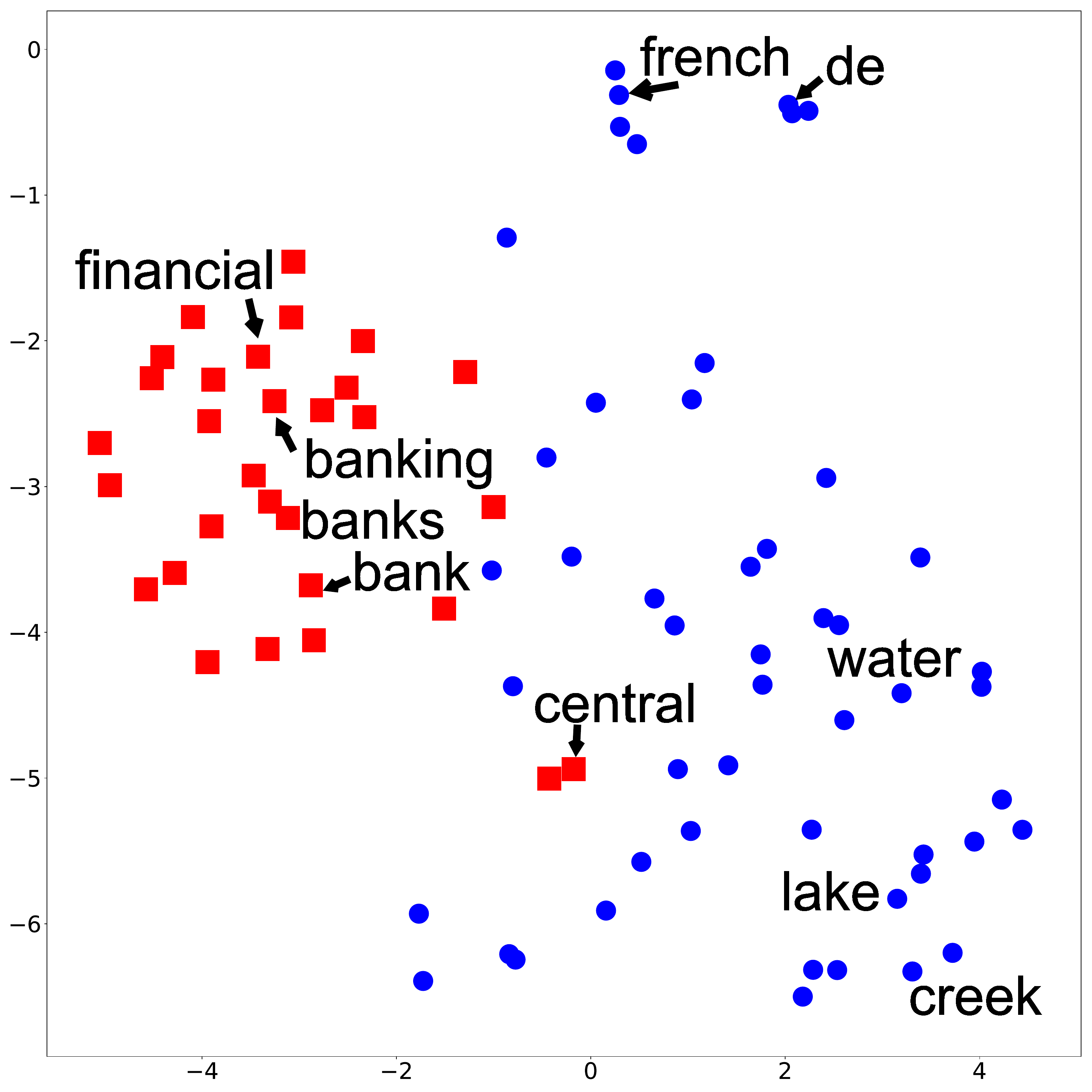}
		\caption{Low-dimensional representation of words and their topic assignments using GLDA}
		\label{fig:GLDAlow}
\end{figure}

\begin{figure}
	\centering
	\includegraphics[width=0.85\linewidth]{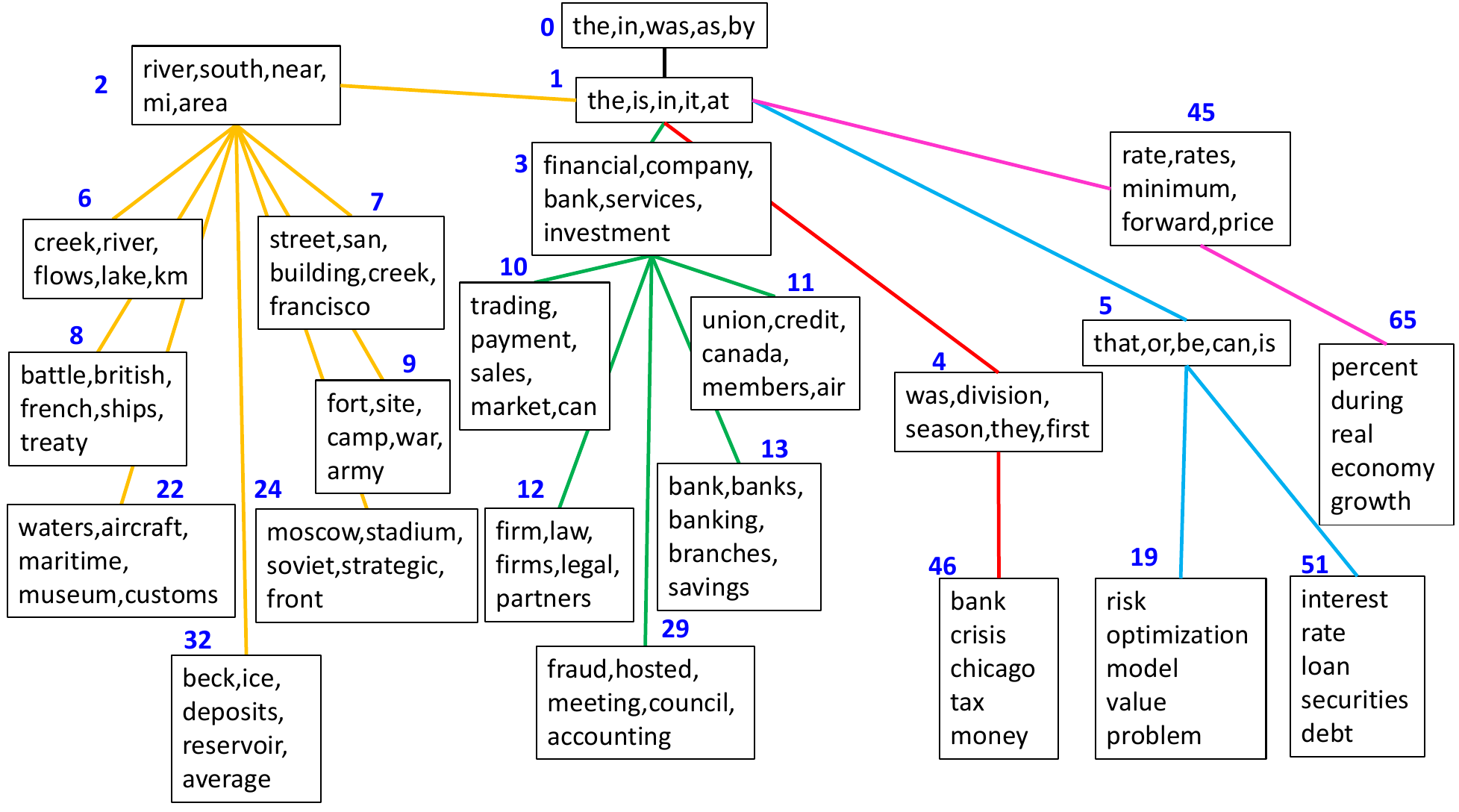}
	\caption{Partial depiction of the topic hierarchy estimated using hLDA}
	\label{hLDAtopicsresult}
\end{figure}

%\begin{figure*}
%	\centering
%	\begin{minipage}{.35\textwidth}
%		\centering
%		\includegraphics[width=0.82\linewidth]{wiki_glove_bank_glda-crop.pdf}
%		\caption{Low-dimensional representation of words and their topic assignments using GLDA}
%		\label{fig:GLDAlow}
%	\end{minipage}%
%	\begin{minipage}{.62\textwidth}
%		\centering
%		\includegraphics[width=0.85\linewidth]{hlda_ncrp_wiki_part-crop.pdf}
%		\caption{Partial depiction of the topic hierarchy estimated using hLDA}
%		\label{hLDAtopicsresult}
%	\end{minipage}
%\end{figure*}

\section{Experiments}
\subsection{Datasets}\label{theway_to_create_datasets}

We conducted experiments using three open datasets, which were all included in our source code.  One of the datasets (i.e., Wikipedia) was assembled particularly for the bank polysemy capturing task.  We summarize the datasets below.
\begin{itemize}
	\item The Wikipedia dataset, abbreviated as Wiki in the table, is a dataset particularly assembled for the bank polysemy capturing task.  The corpus was created from DBpedia-2016 long abstract data~\cite{DBpedia}.  Each long abstract in the DBpedia dataset has several labels that are attached to classify each article.  We focused on the following six categories: ``Rivers,'' ``Banks/Financial,'' ``Military,'' ``Law,'' ``Mathematical,'' and ``Football.'' We sampled evenly from these categories to create a corpus of 6,000, of which 5,000 were used for training and 1,000 for testing.  The main feature of this dataset is the inclusion of the ``Rivers'' and ``Banks/Financial'' categories.  By randomly sampling from these categories, we created a corpus that used ``bank'' both as a financial institution and a steep place near a river.  We used words that appeared more than 50 times in the corpus, and did not remove stop words, as in hLDA~\cite{Blei2010}.  We further focused on words that appeared in all the pre-trained word embeddings described below.
	\item Amazon review data is a dataset of gathered ratings and review information \cite{McAuley2015}\footnote{The entire dataset is available at \url{http://jmcauley.ucsd.edu/data/amazon/}}.  We sampled evenly from the following five categories: ``Electronics,'' ``Video Games,'' ``Home and Kitchen,''
	``Sports and Outdoors,'' and ``Movies and TV,'' and created a corpus of 6,000, of which 5,000 were used for training and 1,000 for testing.  The other settings were the same as above.
	\item Reuters data is a news dataset web-scraped from Reuters news.  We collected 6,000 news stories during the period Jan 2016 to Feb 2016, of which 5,000 were used for training and 1,000 for testing.  The other settings were the same as above.
\end{itemize}
% (i.e., three corpora times three word embeddings)
For pre-trained word embedding vectors, we used the GloVe (50 dimension)~\cite{Pennington2014}, word2vec (300 dimension)~\cite{Mikolov2013}, and fasttext (300 dimension)~\cite{Bojanowski2016} word embedding vectors.  Hence, in total, we had nine settings for models using word embeddings.

\begin{figure}
		\centering
		\includegraphics[width=0.5\linewidth]{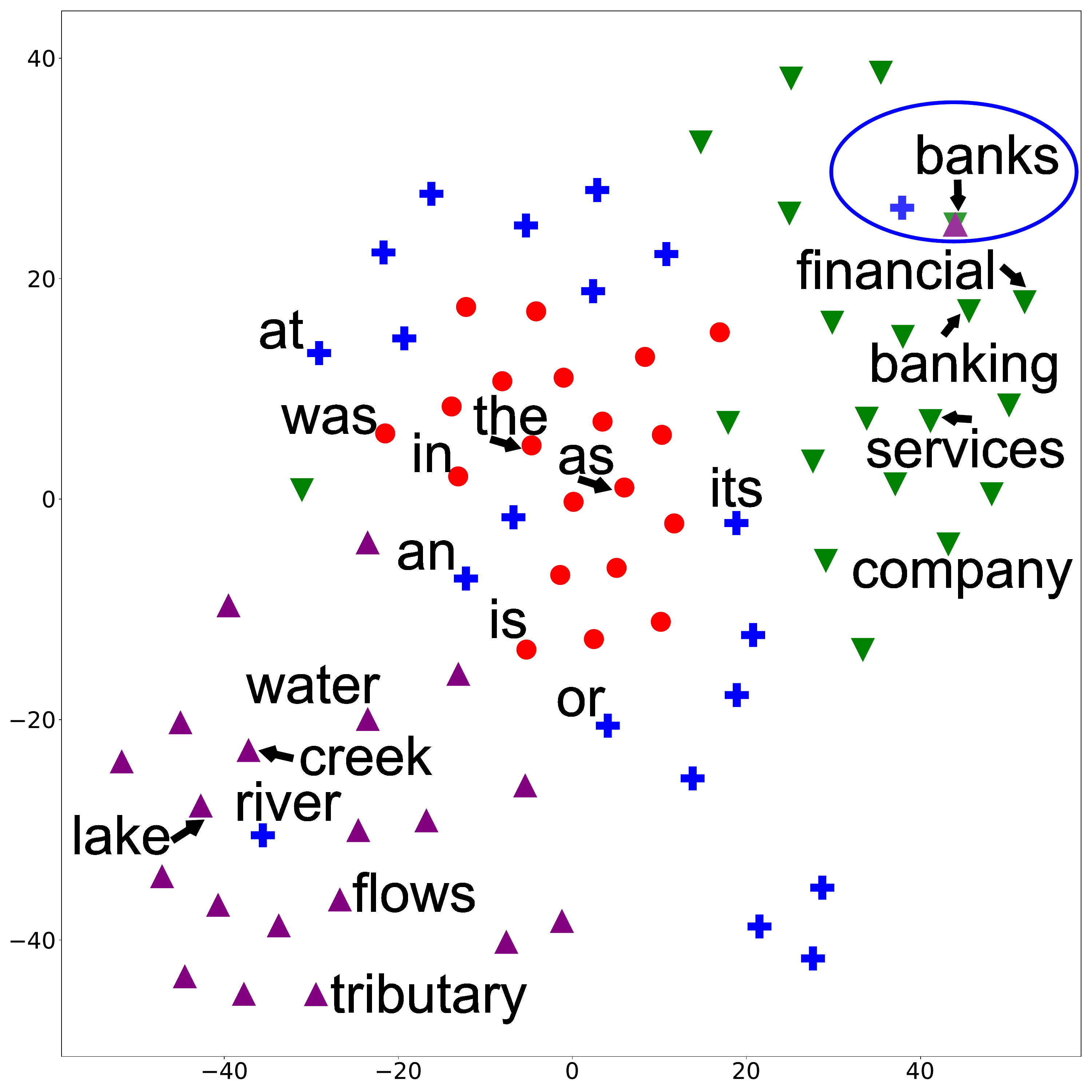}
		\caption{Low-dimensional representation of words and their topic assignments using GhLDA}
		\label{fig:GhLDAlow}
\end{figure}

\begin{figure}
	\centering
	\includegraphics[width=0.85\linewidth]{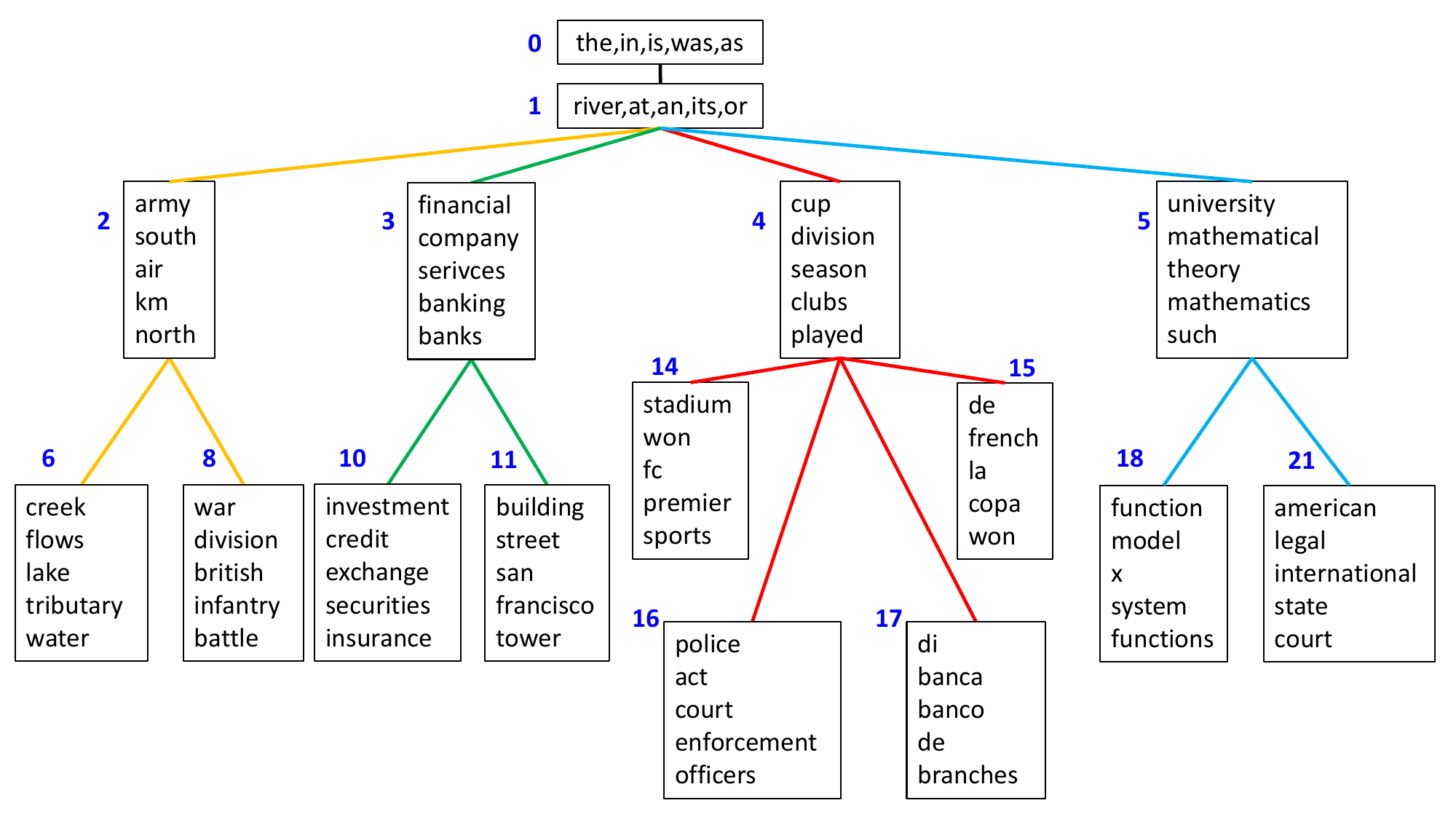}
	\caption{Topic hierarchy estimated using GhLDA}
	\label{fig:GhLDAtopics}
\end{figure}

%\begin{figure*}
%	\centering
%	\begin{minipage}{.35\textwidth}
%		\centering
%		\includegraphics[width=0.8\linewidth]{wiki_glove_bank_ghlda.pdf}
%		\caption{Low-dimensional representation of words and their topic assignments using GhLDA}
%		\label{fig:GhLDAlow}
%	\end{minipage}
%	\begin{minipage}{.64\textwidth}
%		\centering
%		\includegraphics[width=0.85\linewidth]{ghlda_ncrp_0105-5.pdf}
%		\caption{Topic hierarchy estimated using GhLDA}
%		\label{fig:GhLDAtopics}
%	\end{minipage}%
%\end{figure*}

\subsection{Settings}

We compared GhLDA with LDA, hLDA~\cite{Blei2010}, GLDA~\cite{Das2015}, and CGTM~\cite{Xun2017}.  For the topic coherence and predictive held-out likelihood experiments, the number of topics for LDA, GLDA, and CGTM was fixed to 40. For our qualitative analysis, we also considered the case of 20 topics.

The hyperparameters that governed the topic distributions were set to $\alpha=0.1,\beta=0.1$ for LDA, and $v=0.1,\kappa=0.1,\Psi_{glove}=50*I,\Psi_{word2vec}=40*I,\Psi_{fasttext}=20*I$ for GLDA and CGTM, where $I$ denotes an identity matrix.  We ran the sampler for 50 epochs for these models, where one epoch was equal to sampling all the word positions in the corpus once. The hyperparameters controlling $GEM$ and $nCRP$ were set to $m=0.5,b=100,\gamma=0.1$ similar to~\cite{Blei2010}.  The initial tree structure of hLDA and GhLDA was set to $[1,1,4,4]$, where each number corresponds to the number of branches at each level.  In hLDA, $\eta$ was set to vary among the levels as $[2,1,0.5,0.25]$.  A similar strategy was used in GhLDA, where we adjusted $\Psi$ to vary among the levels in the ratio $[1,0.8,0.6,0.4]$, where the top level was identical to GLDA.  We truncated the tree at level four, as in \cite{Blei2010}.  For GhLDA, we further ran the sampler without adding any leaves for five epochs.  For the initial level assignments, half of the assignments were chosen by dividing the cumulative distribution function of word frequency into four segments and assigning from top to bottom according to the segments. The other half was chosen randomly. These additional steps were performed to stabilize the learning of the Gaussian mixture components.  We ran the sampler for 100 epochs.

\subsection{Capturing Polysemy}

We compare the models' ability to capture polysemy, paying particular attention to the term ``bank(s)," using the Wikipedia dataset.  We use GloVe as a case study; the other word embeddings provide similar results.  First, as shown in Table.~\ref{Table: GLDA}, in topics trained using GLDA with $K=20$, topic 10 included terms related to finance, such as ``financial," ``banking," and ``central," and topic 13 contained terms related to the river, such as ``creek," ``lake," and ``water."  However, not a single ``bank" or ``banks" that appeared in the corpus was assigned to the river topic (i.e., topic 13), and all these words were assigned to the finance topic (i.e., topic 10).  Similar observations were made, even when $K$ was increased to 40.  In this case, we could see terms related to finance, such as ``financial," ``market," and ``management," in topic 21, and terms related to river, such as ``river," ``creek," and ``flows," in topic 0.  However, topic 0 also contained financial terms, such as `` investment," ``credit," and ``exchange;" hence, the topic was inappropriately mixed.  This observation implies that although increasing the number of topics makes the constraint on mutual exclusivity to soften; it does not improve the ability to capture polysemy.  Similar observations hold for CGTM as well.

By contrast, GhLDA can capture the polysemy of ``bank(s)."  As is shown in Fig.~\ref{fig:GhLDAtopics}, path [0-1-2-6] is related to the river and path [0-1-3-10] is related to finance.  Although all uses of ``bank" in the Wikipedia dataset were assigned to topic 1 because of the high frequency of the word, the meaning could be discerned from the path assignment.  Moreover, ``banks" in the dataset were assigned to the correct topic (i.e., either topic 3 or 6) in terms of the label of the documents (i.e., we utilized ``Rivers" and ``Banks/Financial" categories explained in the dataset section). Hence, we observe that GhLDA can distinguish ``bank(s)" polysemy.

The inability to capture polysemy in GLDA is further illustrated using low-dimensional representations.  Fig.~\ref{fig:GLDAlow} shows each word's assignment of topics 10 and 13 in addition to their two-dimensional representation using T-sne~\cite{maaten2008}.  We can see that ``bank(s)" is far apart from terms related to the river, and ``bank(s)" is never assigned to the topic about rivers.  As a comparison, Fig.~\ref{fig:GhLDAlow} shows the two-dimensional representation of GhLDA.  We can see that although ``banks" is surrounded by terms that relate to finance, ``banks," which is located in the upper right of the figure, is also assigned to a path that refers to rivers, showing that GhLDA can capture polysemy. Similar observations hold for other words, such as ``law" and ``order" (i.e., paths [0-1-5-15] and [0-1-5-21]), as well.

We further note that, because our corpus does not exclude highly frequent terms as in ~\cite{Blei2010}, LDA cannot capture polysemy well (i.e., Table.~\ref{Table: GLDA}) because the topics are contaminated with stop words, and it is difficult to distinguish the difference between topics.

\subsection{Comparison with hLDA}\label{comparisonGhLDAandhLDA}

In this section, we mainly focus on the difference between GhLDA and hLDA.  As shown in Fig.~\ref{hLDAtopicsresult}, the main difference can be seen in the hierarchical structure learned between the two models. Whereas the numbers of paths and topics estimated in hLDA are 54 and 83, in GhLDA, they are 10 and 16, respectively, on the Wikipedia datasets, which shows that hLDA tends to have a higher number of paths and topics than GhLDA.  Both GhLDA and hLDA have paths for finance (e.g., [0-1-3-10],[0-1-4-46],[0-1-45-65]) and the river (e.g., [0-1-2-6]), thus capturing the polysemy of words (i.e., Table.~\ref{characteristic}).  However, too many paths in hLDA cause crucial redundancy. For instance, there are seven paths related to finance that sometimes have no apparent distinction between them (e.g., [0,1,45,65] and [0,1,5,51]).  These redundancy hearts the coherency of topics, as we show in the next section.
\begin{table}
	\caption{Topic coherence}
	\label{PMI}
	\centering
	\resizebox{0.6\columnwidth}{!}{
		\begin{tabular}{lrrr}
			\toprule
			Corpus &      Wiki &    Amazon &   Reuters \\
			\midrule
			LDA            & -3.32 & -1.94 & -3.58 \\
			hLDA           & -1.05 & -1.50 & -1.55 \\
			GLDA-GloVe     & -1.13 & -1.65 & {\bf -1.17} \\
			GLDA-word2vec  & -1.75 & -1.92 & -1.80 \\
			GLDA-fasttext  & -1.96 & -1.88 & -2.07 \\
			CGTM-GloVe     & -0.93 & -1.34 & -1.41 \\
			CGTM-word2vec  & -1.63 & -1.76 & -1.85 \\
			CGTM-fasttext  & -1.87 & -1.77 & -1.94 \\
			GhLDA-GloVe    & -0.79 & -1.54 & -1.53 \\
			GhLDA-word2vec & {\bf -0.60} & -1.66 & -1.54 \\
			GhLDA-fasttext & -1.06 & {\bf -1.23} & -2.16 \\
			\bottomrule
	\end{tabular}}
\end{table}
\subsection{Topic Coherence}
We calculated the topic coherence score~\cite{Newman2010, ChangReadTea} using Palmetto~\cite{TopicEval2015} to check how coherently each model generates topics.  We computed the average topic coherence score using the basic pointwise mutual information (PMI) measure, focusing on the top 10 words.  Table~\ref{PMI} summarizes the results.  First, we see that compared to LDA and hLDA models, using word embedding tends to outperform the no word embedding counterparts.  Among the models that use word embeddings, GhLDA was the best model, except on the Reuters dataset.

However, the topics learned from the Reuters dataset using GhLDA were not at all worse than the GLDA and CGTM counterpart.  For instance, in GhLDA-word2vec, there were topics such as ``trump, republican, coal, party, workers, house, debate, school, bill, bankruptcy," which indicate the news topic that Trump made a promise to coal miners during his campaign, and ``vehicles, water, vw, flint, safety, cars, emissions, volkswagen, detroit, filed," which indicate the news topics of Volkswagen's diesel cars and the tap water problem of Flint.  Although these news topics were widely reported during the period in which the news dataset was collected, the PMIs of the topics were -1.36 and -2.79, respectively, which shows the limitations of Palmetto for evaluating new combinations of words correctly.

Furthermore, even though they connect to real word news, neither Trump nor Volkswagen appeared in the top 15 words of the 40 topics learned from GLDA-word2vec and CGTM-word2vec.  Topics in GLDA were much general, such as ``rate, dollar, assets, buy, goal, drop" and ``government, end, federal, chinese, countries," which do not take into the word co-occurrence patterns of the corpus that we wish to analyze.  As the Reuters examples suggest, even when the underlying word embedding is not in line with the corpus, the added flexibility of our model identifies critical topics that both GLDA and CGTM fail to identify.  This observation further highlights the benefit of our model.
\subsection{Quantitative Comparison}
We further used the predictive held-out likelihood to quantitatively compare our models, as in \cite{Blei2010}.  We evaluated the probability of the held-out dataset using the 1,000 test documents described in the dataset section.  \cite{Blei2010} used the harmonic mean~\cite{Kass1995} to evaluate the held-out likelihood.  However, \cite{Wallach2009, Buntine2009} showed that the harmonic mean method is biased. Hence, we used the left-to-right sequential sampler~\cite{Buntine2009},  which estimates the quantity:
\begin{equation}
p(w_{d,1:N_{d}}|\alpha,\theta) = \prod_{n \leq N_{d}}\Sigma_{k_{n}}\theta_{k_{n},w_{d,n}} p(k_{n}|k_{1:n-1},\alpha,\theta),
\end{equation}
\noindent and to make a fair comparison of the models, we evaluated $\theta_{k_{n},j_{n}}$ for all the models using the topic assignments derived from each model and assessed the likelihood.  Table~\ref{heldout} summarizes the results. First, models that use word embedding exhibited better results.  Second, we see that CGTM beat GLDA significantly because of the additional correlation structure.  Even without using word embedding, hLDA further beat the other two models by a large margin.  However, GhLDA seemed to be the best model, outperforming all the other models.
\begin{table}
	\caption{Predictive held-out likelihood}
	\label{heldout}
	\centering
	\resizebox{0.6\columnwidth}{!}{
		\begin{tabular}{llll}
			\toprule
			Corpus & Wiki & Amazon & Reuters \\
			\midrule
			LDA &  -1087.6 &  -1218.9 &  -1926.0\\
			hLDA &     -838.9 &  -1073.0 &  -1676.8\\
			GLDA-glove &   -1016.9 &  -1105.5 &  -1757.3 \\
			GLDA-word2vec &   -1019.4 &  -1105.4 &  -1758.4 \\
			GLDA-fasttext &   -1021.6 &  -1105.4 &  -1452.9 \\
			CGTM-glove &     -946.7 &  -1046.8 &  -1667.6 \\
			CGTM-word2vec &     -948.1 &  -1029.8 &  -1667.1\\
			CGTM-fasttext &  -943.3 &  -1049.3 &  -1665.7\\
			GhLDA-glove &     \textbf{-558.7} &     \textbf{-659.1} &  -1127.0 \\
			GhLDA-word2vec &     -577.9 &     -664.7 &  \textbf{-1078.0} \\
			GhLDA-fasttext &     -578.9 &     -660.2&  -1079.4\\
			\bottomrule
	\end{tabular}}
\end{table}

\section{Conclusion}

In this paper, we proposed Gaussian hLDA, which significantly improves the capture of polysemy compared with GLDA and CGTM.  Our model learns the underlying topic distribution and hierarchical structure among topics simultaneously, which can be further used to understand the correlation among topics.  The added flexibility of our model does not necessarily increase the time complexity compared with GLDA and CGTM, which makes our model a good competitor to GLDA.  We demonstrated the validity of our approach using three real-world datasets.  

%We showed that our model improves both in terms of held--out predictive likelihood and topic coherence compared to the state--of--the--art models.

%\section{Acknowledgment}

%We would like to thank Tsutomu Watanabe, Takaaki Ohnishi, Hiroshi Iyetomi, Daisuke Inoue, Taichi Kiwaki, and Kohei Miyaguchi for vigorous discussions.  R.H. was supported by ACT-i, JST.  T.M. was supported by Grant-in-Aid for Young Scientists (A) \#16H05904, JSPS.  Financial support from CARF at the University of Tokyo is also acknowledged.  We thank Kim Moravec, Ph.D., and Maxine Garcia, Ph.D., from Edanz Group for editing a draft of this manuscript. 		

% Bibliography
%\bibliographystyle{ACM-Reference-Format}
\bibliographystyle{plain}
\bibliography{Manu_Bib}

\end{document}